\documentclass[a4paper,conference]{IEEEtran}
\usepackage{mathtools}

\usepackage{cite}

  \usepackage[pdftex]{graphicx}
\usepackage{amsmath}
\usepackage{array}
\usepackage{multirow}
\begin{document}
%
\title{Weakly Supervised Medical Image Segmentation With Soft Labels and Noise Robust Loss}



%
\author{\IEEEauthorblockN{Banafshe Felfeliyan\IEEEauthorrefmark{1}\IEEEauthorrefmark{2},
Abhilash Hareendranathan\IEEEauthorrefmark{3},
Gregor Kuntze\IEEEauthorrefmark{2},
Stephanie Wichuk\IEEEauthorrefmark{3},\\
Nils D. Forkert\IEEEauthorrefmark{1},
Jacob L. Jaremko\IEEEauthorrefmark{3}\IEEEauthorrefmark{4}, and
Janet L. Ronsky\IEEEauthorrefmark{1}\IEEEauthorrefmark{2}\IEEEauthorrefmark{5}}
\IEEEauthorblockA{\IEEEauthorrefmark{1}Department of Biomedical Engineering, University of Calgary, Calgary, Alberta, Canada}
\IEEEauthorblockA{\IEEEauthorrefmark{2}McCaig Institute for Bone and Joint Health University of Calgary, Calgary, Alberta, Canada}
\IEEEauthorblockA{\IEEEauthorrefmark{3}Department of Radiology \& Diagnostic Imaging, University of Alberta, Edmonton, Alberta, Canada}
\IEEEauthorblockA{\IEEEauthorrefmark{4}Alberta Machine Intelligence Institute (AMII), University of Alberta, Edmonton, Alberta, Canada}
\IEEEauthorblockA{\IEEEauthorrefmark{5}Mechanical and Manufacturing Engineering, University of Calgary, Calgary, Canada}}


\maketitle

\begin{abstract}
Recent advances in deep learning algorithms have led to significant benefits for solving many medical image analysis problems. Training deep learning models commonly requires large datasets with expert-labeled annotations. However, acquiring expert-labeled annotation is not only expensive but also
is subjective, error-prone, and inter-/intra- observer variability introduces noise to labels. This is particularly a problem when using deep learning models for segmenting medical images due to the ambiguous anatomical boundaries. Image-based medical diagnosis tools using deep learning models trained with incorrect segmentation labels can lead to false diagnoses and treatment suggestions. Multi-rater annotations might be better suited to train deep learning models with small training sets compared to
single-rater annotations. The aim of this paper was to develop and evaluate a method to generate probabilistic labels based on multi-rater annotations and anatomical knowledge of the lesion
features in MRI and a method to train segmentation models using probabilistic labels using normalized active-passive loss as a "noise-tolerant loss" function. The model was evaluated by comparing it to binary ground truth for 17 knees MRI scans for clinical segmentation and detection of bone marrow lesions (BML). The proposed method successfully improved precision 14, recall 22, and Dice score 8 percent compared to a binary cross-entropy loss function. Overall, the results of this work suggest that the proposed normalized active-passive loss using soft labels successfully mitigated the effects of noisy labels.
\end{abstract}


%
\IEEEpeerreviewmaketitle

\section{Introduction}
Automatic or semi-automatic clinical features and biomarker measurements based on deep learning (DL) are useful for longitudinal assessment of  medical images. 
DL-based techniques using Convolutional Neural Networks (CNN) have shown great success in tissue and pathology detection and segmentation \cite{felfeliyan2022improved,pedoia20193d}.
The accuracy of deep learning methods is highly dependent on the quality of the training data and corresponding ground truth labels. For sensitive applications like medical image segmentation, it is particularly important to use anatomically accurate labels for training of DL models. 
However, acquiring “ground truth” (GT) labels in the medical domain can be challenging because pixel-wise labeling is expensive, subjective, and error-prone. Inter-reader variability combined with the presence of ambiguous anatomical boundaries between tissues (\textit{e.g.}, due to partial volume effects) makes labels uncertain \cite{gros2021softseg}. Furthermore, the presence of lesions and pathologies in unexpected locations increases the chance of inattentional blindness \cite{busby2018bias}. Consequently, medical image datasets are likely to contain sub-optimal, inaccurate and noisy labels \cite{KARIMI2020101759}.\par
A DL model trained on weak/noisy annotations may have biases and overfit to incorrect annotations \cite{zhang2021understanding}. Therefore, different approaches have been investigated to reduce the effects of noise and imperfect labels in medical image analysis, including label smoothing and label correction \cite{islam2021spatially,to2022coarse}. Despite the efforts to develop noise-resistant learning approaches, many aspects have remained unexplored, particularly for medical image segmentation tasks. Therefore, for medical image segmentation task, there is a high demand for robust and reliable methods for training DL models based on noisy and suboptimal labels (\textit{e.g.}, annotations that are not pixel-wise or partial labels).\par In this work, we proposed to train an instance segmentation model with soft labels obtained from noisy/partial region of interest (ROI) labels using a noise resistance loss function. We assume that partial ROI labels as highly noisy labels and attempt to perform weakly supervised instance segmentation under this assumption. Our key contributions are: 
\begin{enumerate}
  \item Exploration of training of an improved version of MaskRCNN \cite{felfeliyan2022improved} using probabilistic partial labels.
  \item Creation of probabilistic labels based on multi-rater scoring, MRI spatial redundancy, and tissue/lesion characteristics, as well as considering image contextual information.
  \item Proposal of combining a noise-tolerant loss function (active passive loss) and soft ground truth for training DL models.
\end{enumerate}
\par As a practical application, we focus on bone marrow lesion (BML) detection and segmentation, which is one of the inflammatory components of osteoarthritis (OA).Accurate quantification of features related to OA inflammation can provide a basis for effective clinical management and a target for therapy \cite{jaremko2017preliminary}. This task is challenging as BMLs do not have distinctive edges and it is challenging to create binary labels. Furthermore, it is hard to generate precise clean annotations for BMLs, since they may appear in multiple locations, and humans are susceptible to inattentional blindness (missing BMLs in plain sight) \cite{busby2018bias}.


\section{Related Work}
\subsection{Learning with noisy labels}
To mitigate the effect of noisy labels, different approaches like label correction, noise-robust loss, robust regularization, and loss correction have been deployed, which are briefly described in the following.

\subsubsection{Label correction}
Label correction aims to improve the quality of raw labels. Different methods have been proposed to estimate label noise distribution, correct corrupted labels \cite{yi2021learning}, or separate the noise from the data using properties of learned representations \cite{liu2022robust,kim2021fine}. Even though the label correction methods are effective, they usually require additional precise annotated data or an expensive process of noise detection and correction \cite{ma2020normalized,ibrahim2020semi}.
\subsubsection{Noise-robust loss}
One approach for noise-robust learning is using loss functions that are inherently noise-tolerant
or losses that created by using regularization terms or modifying well-known loss functions to make them noise-tolerant.\par Ghosh et al. theoretically proved that symmetric losses perform significantly better in case of learning with noisy labels \cite{ghosh2017robust}. 
The Mean Absolute Error (MAE) loss is symmetric and noise-tolerant for uniform noise and class-conditional label noise and satisfies the symmetry condition. In contrast to that, cross-entropy (CE) is not symmetric and does not perform well in the presence of noise \cite{ghosh2017robust}. Therefore, training with CE in addition to complementary losses robust to noise to achieve learning sufficiency and robustness has been suggested \cite{wang2019symmetric,wang2021learning}. Wang et al. demonstrated that the Reverse Cross-Entropy (RCE) \cite{ghosh2017robust} loss, which is robust to label noise, can be used as the complementary robust loss term \cite{wang2019symmetric}. They proposed the so-called Symmetric Cross-Entropy (SCE) loss, which is defined as $l_{SCE}=\alpha l_{ce} +\beta l_{Rrce}$, where $l_{ce}$ is the CE loss and $l_{Rrce}$ is the RCE loss. Using the SCE idea, Ma et al. recently addressed the  learning under label noise by characterizing existing loss functions into two types Active vs. Passive \cite{ma2020normalized}. Based on their characterization the active loss is a loss that maximizes $p(k=y|x)$ and the passive loss is a loss that minimizes $p(k\neq y|x)$. Then they proposed the Active and Passive Loss (APL), which combines an active loss and a passive loss in order to maximize the probability of belonging to a given class and minimize the likelihood of belonging to another class \cite{ma2020normalized}. The APL loss was shown to be  noise-tolerant if both active loss and passive loss have been chosen from  noise-tolerant losses \cite{ma2020normalized}. In the same work, Ma et al. proved that any loss can be noise-tolerant if normalization is applied to it \cite{ma2020normalized}. Their results showed that APL addresses the underfitting problem and can leverage both robustness and convergence advantages.
\subsubsection{Robust regularization}
Regularization methods have been widely used to increase robustness of deep learning models against label noise by preventing overfitting.
These methods perform well in the presence of moderate noise, and are mostly used in combination with other techniques \cite{song2022learning}. Different regularization techniques include explicit regularization like weight decay and dropout or implicit regularization like label smoothing.
\subsubsection{Loss correction}
Most of loss correction methods estimate noise-transition matrices to adjust the labels during training. These techniques aim to minimize global risk with respect to the probability distribution. The backward and forward correction involves modifying two losses based on the noise transition matrix \cite{patrini2017making}. Some loss correction methods assume the availability of trusted data with clean labels for validation and to estimate the matrix of corruption probabilities \cite{hendrycks2018using} or they may require complex modifications to the training process.\par
Recently, Lukasik et al. \cite{lukasik2020does} have shown that label smoothing is related to loss-correction techniques and is effective in coping with label noise by assuming smoothing as a form of shrinkage regularization. To mitigate noise, Label smoothing may require a simple modification to the training process and does not require additional labels and offers loss correction benefits.

\subsection{Label smoothing and soft labels}
Traditionally, the label ($y$) of each pixel is encoded binary (or as a one-hot vector). In binary labeling, one instance belongs to either one or the other class. This hard labeling assigns all probability mass to one class, resulting in large differences between the largest logit and the rest of the logits in networks using the sigmoid (or soft-max) activation function in the output layer \cite{vega2021sample}. Consequently, in applications such as medical imaging, in case of partial volume effects or disagreement between readers, hard labels may cause overfitting and reduce the adaptability of the network in these situations \cite{vega2021sample}. In contrast, soft labels encode the label of each instance as a real value probability, whose k-th entry represents $p(Y=k|X=x)\in[0,1]$. For example, the soft label $x1=[0.3,0.7]$ indicates that $p(Y=1|X=x_i)=0.7$, whereas hard labels only have 0/1 values. Soft labels can provide additional information to the learning algorithm, which was shown to reduce the number of instances required to train a model \cite{vega2021sample}. \par
Label smoothing techniques can be considered as utilization of probabilistic labels (soft labels). These approaches have been shown to give a better generalization, faster learning speed, better calibration, and mitigation of network over-confidence \cite{muller2019does}. Label smoothing has been deployed and examined in different applications including image classification \cite{vega2021sample}, model network uncertainty \cite{silva2021using}, and recognition \cite{NIPS2017_3f5ee243}. Probabilistic labels have been used to mitigate reader variability and human assessment noise for classification problems \cite{xue2017efficient}.\par
While methods based on uniform label smoothing techniques may improve the network calibration, they do not necessarily improve classification accuracy. As a solution, Vega et al. proposed to compute probabilistic labels from relevant features$(Z(X))$ in the raw images $(X)$ for a classification task \cite{vega2021sample}. In that work, a method is trained to estimate $p(Y|Z(X))$, which is used as a probabilistic labels’ classification task. This probabilistic labeling approach was shown to provide better calibration and improved classification accuracy than uniform label smoothing.
\subsubsection{Soft labels for medical image segmentation}
Smoothing labels and probabilistic labels have only been investigated recently in segmentation tasks. This is because earlier developed methods of label smoothing were originally proposed for image classification, in which the hard labels were flattened by assigning a uniform distribution overall to all classes to prevent model overconfidence. However, this is likely problematic in segmentation tasks, since this approach assigns a probability greater than zero to pixels, even those that one can be confident about not belonging to a certain class (\textit{e.g.}, background outside the body). \par
Islam et al. addressed this problem by proposing a Spatially Varying Label Smoothing (SVLS) to capture expert annotation uncertainty \cite{islam2021spatially}. SVLS considers the likelihood with neighboring pixels for determining the probability of each class. Li et al. proposed super-pixel label softening to encounter descriptive contextual information for soft labeling. Using super-pixel soft labels and KL (Kullback-Leibler divergence) loss improved Dice coefficient and volumetric similarity \cite{li2020superpixel}. \par
The high uncertainty in defining lesion borders was investigated by Kats et al. who developed a model that uses soft labels (generated by morphological dilation of binary labels) with the soft-Dice loss to segment multiple-sclerosis lesions in MRI data to account for the high uncertainty in defining lesion borders \cite{kats2019soft}. However, further analysis revealed that the soft-Dice may introduce a volumetric bias for tasks with high inherent uncertainty \cite{bertels2019optimization}. Gross et al. used soft labels obtained through data augmentation for the segmentation task \cite{gros2021softseg}. They assumed that using soft labels is analogous to a regression problem, and they used the NormReLU activation and a regression loss function for medical image segmentation \cite{gros2021softseg}. However, using NormReLU as the last layer has two main drawbacks. First, it is not highly effective when the maximum is an outlier. Second, it only uses one image to normalize ReLU, which may cause the algorithm to not converge to a good solution since a single image may not be a good representation of the entire data set distributions. \par
In some studies, soft labels were generated by fusing multiple manual annotations. One of the best methods to obtain a consensus mask from multi-reader annotation is the Therefore, Kats et al. proposed a soft version of the STAPLE algorithm \cite{kats2019softSTAPLE}, which showed superior results compared to morphological dilation used in the aforementioned soft Dice loss approach \cite{kats2019soft}. 

\section{Method}
Ma et al. \cite{ma2020normalized} and Lukasik et al. \cite{lukasik2020does} showed that a normalized loss function and label smoothing are both effective ways to mitigate noise effects. In this paper, we are investigating whether the combination of label smoothing, and a normalized loss function leads to quantitative benefits.
\subsection{Problem Formulation}
The aim of supervised learning for classification problems is to learn the function $f(x;\theta)$. This function maps the input $x$ to the output $y$ using a deep neural network parametrized by $\theta$. $f$ approximates the underlying conditional distribution $p(y|x;\theta)$ to minimize the loss function. We can define a strongly labeled dataset with correct annotation as $D_{S}=\{(x,y_S)_{m}\}_{1\leq m\leq |D|}$, where $y_S \in 0,1$. In addition, the weakly soft labeled set (noisy labeled) can be defined as $D_{W}=\{(x,\bar{y})_{n}\}_{1\leq n\leq |D|}$, $\bar{y}\in [0,1]$, where $x^{(i)}\in \Re^{n_{x}}$
 denotes input MRI image (feature space), $y_{s}^{(i)}\in \Re^{m_{y}}$ is the distribution correct strong labels, and $\bar{y}^{(i)}\in \Re^{n_{y}}$ is the distribution of observed labels (weak soft labels), $K$ the number of segmentation classes, and $\hat{y}_{i}=f(x_i)$ the output of the model given pixel $i$. Consequently, $p(\hat{y}_{i,k}=k|x_i;\theta)$ is the probability that pixel $i$ is assigned to class $k\in K$   (denoted as $\hat{p}_{i,k}$). \par
 The problem with weak labels (poor, noisy, and partial) is that the probability distribution of the observed label is not equal to the correct labels $P(\bar{Y}_{W}|X)\neq P(Y_{S}|X)$, which is typically caused by (a) partial instance segmentations or (b) missing object instances in the observed labels. \par
 In classification problems, the aim is to minimize the risk of $f$, defined as $R(f)=E_{p(x,y)}[l(Y,f(X))]$ where $l(Y,f(X))$ is the loss function. The goal of classification with weak labels continues to minimize the classification risk, defined as $\bar{R}(f)=E_{p(X,\bar{Y})}[\bar{l}(\bar{Y},f(X))]$, where $\bar{l}(\bar{Y},f(X))$ is a proper loss function for learning noisy labels. Using a noise-tolerant loss, assuming that $f^{*}$ that minimizes $\hat{R}(f)$ can be determined, would be a global minimizer of $R(f)$ as well. In contrast, DL models using loss functions without robustness to noisy labels tend to memorize the noisy samples to minimize the $\hat{R}(f)$ risk. Therefore, in this paper, we are using soft labels and a noise-robust loss function to combat weak label problem and to determine $f^{*}$ to minimize $\hat{R}(f)$, which is also a global minimizer of $R(f)$.
 \subsection{Model}
 We used the IMaskRCNN \cite{felfeliyan2022improved} as the baseline deep learning model for training and evaluation of the proposed extensions. The IMaskRCNN model is an improved version of the well-known instance segmentation model Mask RCNN \cite{8372616} that improves the segmentation accuracy around object boundaries by adding a skip connection and an extra encoder layer to the mask segmentation head (inspired by U-net architecture) \cite{8372616}. Similar to the original Mask RCNN, the IMaskRCNN is constructed from a backbone (ResNet), which is responsible for feature extraction, a region proposal network (RPN) for extracting the ROI bounding box, and two heads: one for mask segmentation (Mask Head), and the other for classifying the extracted bounding boxes (Classification Head). 
 \subsection{Loss}
 Similar to the Mask RCNN, the IMaskRCNN has a multi-task learning loss ($L$) for each sampled ROI, which is the result of the classification loss, the bounding-box loss, and the mask-loss accumulation $L=L_{cls}+L_{bbox}+L_{mask}$.
 \subsubsection{Mask-Loss for pixel-level noise}
 In Mask RCNN, the mask-head is performing binary segmentation of the detected ROI and $L_{mask}$ is only defined on the $k_{th}$ mask (other mask outputs do not contribute to the loss). Therefore, binary CE (BCE) has been chosen as $L_{mask}$ in Mask R-CNN \cite{8372616}. \par
 Assuming the segmentation error to be the major source of error in comparison with the classification error, we aimed to modify $L_{mask}$ to mitigate the effect of pixel-level noise. Therefore, we propose to replace the BCE loss used as Mask-loss with the APL loss \cite{ma2020normalized} and  adapt the APL loss for soft labels. APL loss is constructed from an active loss term and a passive loss term as:
 \begin{equation}     L_{APL}=\alpha.L_{active}+\beta.L_{passive}
 \end{equation}
 where $\alpha$ and $\beta$ are parameters to balance two terms. \par
 Ma et al. \cite{ma2020normalized} have shown that normalized losses guarantee robustness against noise. Thus, we use normalized losses $(L_{norm})$:
 \begin{equation}
 \label{eq:lnorm}
     L_{norm}=\frac{L(f(x),y)}{\sum^{K}_{j=1}(f(x),j)}
 \end{equation}
 In this paper, we considered Normalized BCE (NBCE) and Normalized RCE (NRCE) for the combination of active and passive losses. \par
 The cross-entropy loss for two distributions, $q$ (GT distribution) and $p$ (predicted distribution), is defined as $H(q,p)$ (eq. \ref{eq:lsce}) and RCE is defined as $H(p,q)$ (eq. \ref{eq:lsrce}). By applying the CE and RCE losses to soft labels $(\bar{y})$, it follows that:
 \begin{equation}
 \label{eq:lsce}
     l_{sce}=H(q,p)=-[\bar{y}\log\hat{y}-(1-\bar{y})\log(1-\hat{y})]
 \end{equation}
 
 \begin{equation}
 \label{eq:lsrce}
     l_{srce}=H(p,q)=-[\hat{y}\log\bar{y}-(1-\hat{y})\log(1-\bar{y})]
\end{equation}
and then the normalized soft CE and normalized soft RCE can be defined using eq. \ref{eq:lnorm}.

 As $log0$ is undefined, the following constraint is applied to probabilities. For $H(a,b)=H(a,b^{*})$, where $b^{*}$ is the probability clipped between two values $b^{*}\in[P_{min},1-P_{min}]$ and $P_{min}=1\mathrm{e}{-20}$:

 \section{Experiments and Results}
\subsection{Data and labels}
In this study, we used the publicly available multicenter Osteoarthritis Initiative (OAI, https://nda.nih.gov/oai/) dataset. OAI contains the demographic and imaging information from 4796 subjects aged 45-79 years who underwent annual knee assessments, including MRIs. A total of 126 knee MRI scans (sagittal intermediate-weighted fat suppressed (IWFS) 444$\times$448 imaging matrix, slice thickness 3 mm, field-of-view 159$\times$160 mm) were selected and scored by experts (2 to 7 readers musculoskeletal radiologists and rheumatologists) for the presence of BML (BML; bright spots in bone) at the tibia, femur, and patella using the Inflammation MRI Scoring System (KIMRISS) \cite{jaremko2017validation}. In this work, we used the BML annotations obtained from KIMRISS along with proxy labels of bones (femur and tibia) obtained from automatic segmentation using the IMaskRCNN trained on registered data from our previous work \cite{felfeliyan2022improved}. In the following, we explain how we prepared BML annotations for training.
\subsubsection{Leverage informative labels from rectangular annotations}
\begin{figure}
\begin{centering}
\includegraphics[width=1\columnwidth]{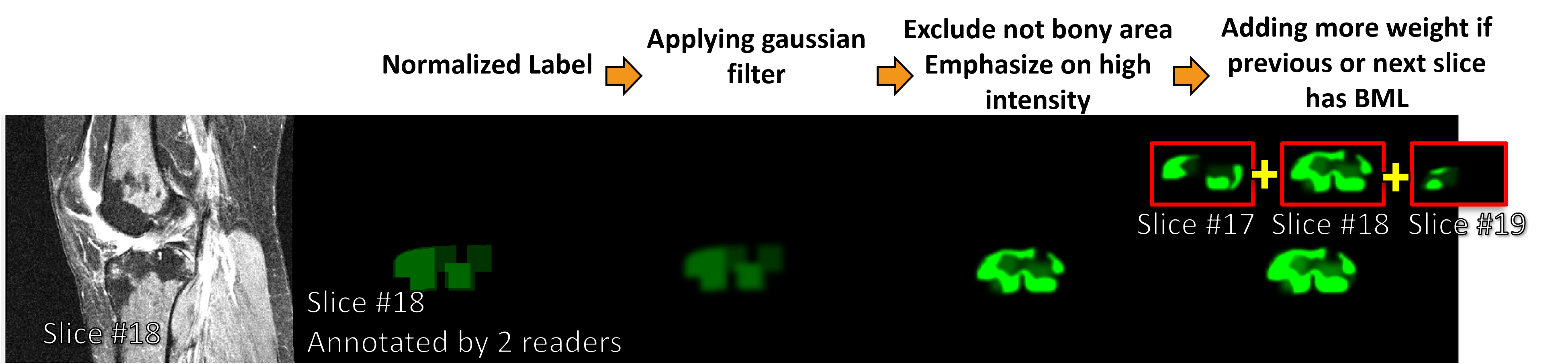}
\par\end{centering}
\caption{\label{fig:methodPipeline}Generating soft pseudo-GT for BML pipeline}
\end{figure}

KIMRISS is a granular semi-quantitative scoring system, which measures inflammation in patients with knee osteoarthritis (OA) \cite{jaremko2017validation}. In the KIMRISS \cite{jaremko2017validation}, by overlaying a transparent grid template on top of the bones (tibia, femur, and patella), the reader specifies regions in slices that contain BML to determine an estimate for BML volume by multiplying granular regions identified as BML \cite{jaremko2017validation}. Using this scoring system and regions specified to have BML it is possible to obtain rectangular ROIs from specified areas on the KIMRISS granular template. In the following, we attempt to first provide cleaner labels and produce soft labels by using scoring results from different raters, spatial redundancy in MRI scans, and BML characteristics in IWFS scans. The following steps were taken to create soft labels from multiple annotations after data cleaning and fixing major errors (Fig. \ref{fig:methodPipeline}).
\begin{enumerate}
    \item Scoring results of raters were aggregated and normalized based on the number of raters who rated the scan.
    \item Bone proxy mask areas of the overlayed grid template outside bone were excluded. 
    \item BML is described as the presence of ill-defined hyperintense areas within trabecular bone on IWFS images. Given this knowledge, we consider areas with greater intensity than other areas of the bone in the trabecular bone more likely to be BML. Therefore, if areas are selected in step 3, we would consider their probability equal to one.
    \item We improved labels by adding labeling information from the previous and next MRI slices, based on the likelihood with neighboring voxels as shown in Fig. \ref{fig:methodPipeline} (i.e., if the previous/next slices of the slice A contains BML, the slice A is more likely to contain BML as well). 
\end{enumerate}
We divided the dataset into 108 scans (from 54 subjects, 1280 slices contain BML) for training, 4 scans (from 2 subjects, 66 slices) for validation, and 17 scans (from 9 subjects, 650 slices) for testing. Furthermore, the validation and test data were segmented manually pixel by pixel by two experts using ITK-snap \cite{hazlett2006h}.
The input images were cropped to 320$\times$320 pixels. To receive more spatial information, we used 2.5D MRI slices (three sequential slices as RGB channels). Furthermore, we have used mirroring as data augmentation to increase number of training data.
\subsection{Implementation}
All models were trained on one NVIDIA V100 GPU for 200 epochs and 200 iterations using an Adam optimizer and a learning rate of 0.001. The complete DL model was implemented in Keras using the TensorFlow 2 backbone. 
\subsection{Evaluation Criteria}
For evaluation, the metrics used included precision $(=TP/(TP+FP))$, recall $(=TP/(TP+FN))$, Intersection of over Union (IoU) $(=(GT\cap Pred)/(GT\cup Pred))$, and average precision (AP), as well as the Dice similarity score $(=  2\times TP / (2\times TP + FP + FN))$ for segmentation. For all of these metrics, higher values indicate better performance.
\subsection{Ablation Study}
Multiple experiments were performed to evaluate the effect of using soft labels and noise resistance loss on BML detection and segmentation with hard and soft labels. Quantitative results of the training with different configurations are summarized in Table \ref{tbl:bmlResults}.

\begin{table*}
\centering
\caption{RESULTS OF DIFFERENT CONFIGURATIONS (FOR BML)}
\label{tbl:bmlResults}
\begin{tabular}{c|cccc|ccccc|c|}
\cline{2-11}
 & \multicolumn{4}{c|}{\textbf{Configuration}} & \multicolumn{5}{c|}{\textbf{Detection}} & \textbf{Seg.} \\ \hline
\multicolumn{1}{|c|}{\multirow{2}{*}{\textbf{Method}}} & \multicolumn{3}{c|}{\textit{Loss weight}} & \multirow{2}{*}{label} & \multicolumn{2}{c|}{AP} & \multicolumn{1}{c|}{\multirow{2}{*}{IoU\%}} & \multicolumn{1}{c|}{\multirow{2}{*}{Rec.}} & \multirow{2}{*}{Prec.} & \multirow{2}{*}{Dice} \\ \cline{2-4} \cline{6-7}
\multicolumn{1}{|c|}{} & \multicolumn{1}{c|}{\textit{\textbf{BCE}}} & \multicolumn{1}{c|}{\textit{\textbf{SCE}}} & \multicolumn{1}{c|}{\textit{\textbf{RCE}}} &  & \multicolumn{1}{c|}{\textit{\textbf{50}}} & \multicolumn{1}{c|}{\textit{\textbf{75}}} & \multicolumn{1}{c|}{} & \multicolumn{1}{c|}{} &  &  \\ \hline
\multicolumn{1}{|c|}{Baseline} & \multicolumn{1}{c|}{1} & \multicolumn{1}{c|}{0} & \multicolumn{1}{c|}{0} & bin & \multicolumn{1}{c|}{12} & \multicolumn{1}{c|}{4} & \multicolumn{1}{c|}{31} & \multicolumn{1}{c|}{0.46} & 0.52 & 0.27 \\ \hline
\multicolumn{1}{|c|}{\multirow{2}{*}{APL binary}} & \multicolumn{1}{c|}{1} & \multicolumn{1}{c|}{0} & \multicolumn{1}{c|}{1} & bin & \multicolumn{1}{c|}{8} & \multicolumn{1}{c|}{3} & \multicolumn{1}{c|}{29} & \multicolumn{1}{c|}{0.43} & 0.47 & 0.21 \\ \cline{2-11} 
\multicolumn{1}{|c|}{} & \multicolumn{1}{c|}{2} & \multicolumn{1}{c|}{0} & \multicolumn{1}{c|}{1} & bin & \multicolumn{1}{c|}{8} & \multicolumn{1}{c|}{1} & \multicolumn{1}{c|}{27} & \multicolumn{1}{c|}{0.26} & 0.40 & 0.20 \\ \hline
\multicolumn{1}{|c|}{Soft Baseline} & \multicolumn{1}{c|}{0} & \multicolumn{1}{c|}{1} & \multicolumn{1}{c|}{0} & soft & \multicolumn{1}{c|}{14} & \multicolumn{1}{c|}{11} & \multicolumn{1}{c|}{37} & \multicolumn{1}{c|}{0.63} & 0.60 & 0.28 \\ \hline
\multicolumn{1}{|c|}{\multirow{4}{*}{APL + soft lbl}} & \multicolumn{1}{c|}{0} & \multicolumn{1}{c|}{1} & \multicolumn{1}{c|}{1} & soft & \multicolumn{1}{c|}{\textbf{19}} & \multicolumn{1}{c|}{\textbf{17}} & \multicolumn{1}{c|}{\textbf{35}} & \multicolumn{1}{c|}{\textbf{0.64}} & \textbf{0.64} & \textbf{0.31} \\ \cline{2-11} 
\multicolumn{1}{|c|}{} & \multicolumn{1}{c|}{0} & \multicolumn{1}{c|}{2} & \multicolumn{1}{c|}{1} & soft & \multicolumn{1}{c|}{12} & \multicolumn{1}{c|}{10} & \multicolumn{1}{c|}{33} & \multicolumn{1}{c|}{0.54} & 0.55 & 0.36 \\ \cline{2-11} 
\multicolumn{1}{|c|}{} & \multicolumn{1}{c|}{1} & \multicolumn{1}{c|}{1} & \multicolumn{1}{c|}{1} & soft & \multicolumn{1}{c|}{\textbf{20}} & \multicolumn{1}{c|}{\textbf{15}} & \multicolumn{1}{c|}{\textbf{38}} & \multicolumn{1}{c|}{\textbf{0.68}} & \textbf{0.66} & \textbf{0.35} \\ \cline{2-11} 
\multicolumn{1}{|c|}{} & \multicolumn{1}{c|}{2} & \multicolumn{1}{c|}{2} & \multicolumn{1}{c|}{1} & soft & \multicolumn{1}{c|}{14} & \multicolumn{1}{c|}{10} & \multicolumn{1}{c|}{30} & \multicolumn{1}{c|}{0.50} & 0.56 & 0.30 \\ \hline
\end{tabular}
\end{table*}

\subsubsection{Soft Labels vs. Hard Labels}
\begin{figure*}
\begin{centering}
\includegraphics[width=1.8\columnwidth]{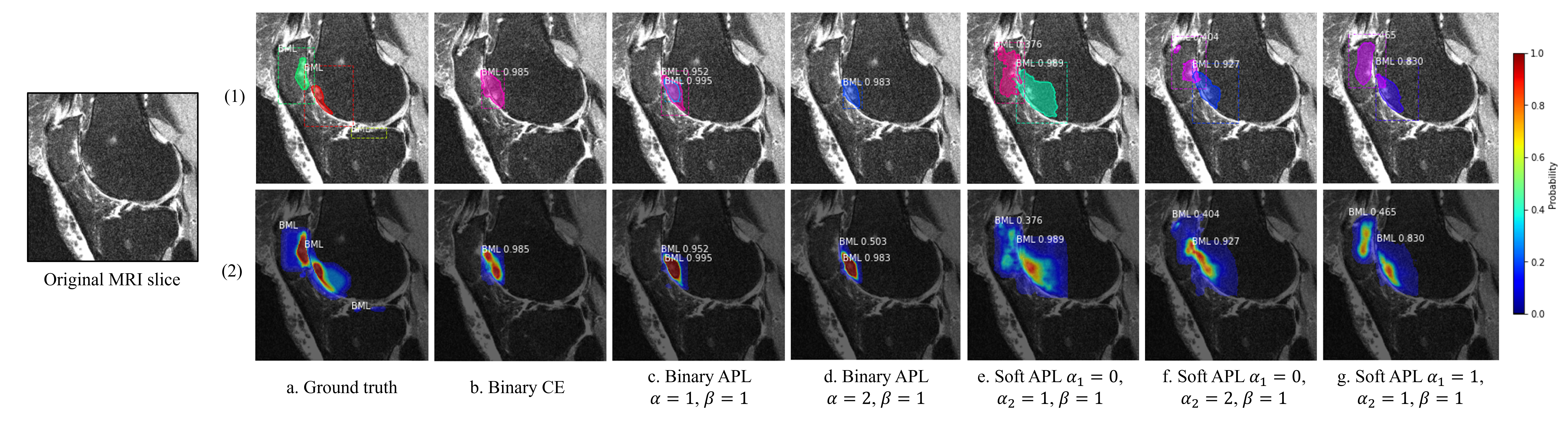}
\par\end{centering}
\caption{\label{fig:result}Results of different configurations for BML detection}
\end{figure*}

We investigated two types of labeling methods, binary labels and soft labels. 
In conventional label smoothing, the labels are smeared based on a $\alpha$ where $p(Y=k|X=x)\in (\alpha,1-\alpha)$. Conventional label smoothing takes binary $(0,1)$ labels and changes its value uniformly. While in our soft labeling approach the $p(Y=k|X=x)$ can have any value between 0 to 1, and label confidence is adjusted using other information like the intensity of pathology, its location and neighboring voxels. The results are shown in TABLE \ref{tbl:bmlResults} demonstrate that using soft labels and a noise resistance loss separately has a positive effect on precision and recall for detecting BML. Comparing the results from BCE (baseline) and soft BCE (soft baseline), shows $4\%$ improvement in recall and $2\%$ improvement in precision and Dice similarity metric. Based on Fig. \ref{fig:result} (a and d), it can be observed that using soft labels is effective in preventing overconfidence (as we expected under noisy labels \cite{lukasik2020does}).
\subsubsection{APL Loss vs. BCE loss} We tested different combinations of active loss $(\alpha_{1}.NBCE+\alpha_{2}.NSCE)$ and passive loss $(\beta.NRCE)$ and compared it to the baseline using the BCE loss. For the active loss, we considered using a combination of the Binary CE and soft CE together. Using APL for binary labels does not seem to be effective. Comparing results from the baseline and APL binary in TABLE \ref{tbl:bmlResults} shows using APL for binary labels  is not improving the metric. Further, as shown in  Fig. \ref{fig:result} (c, d) binary APL also failed to detect patella's BML in the patella.
 
\subsubsection{APL + softLabels} The combination of APL + softLabels has improved the recall and precision $10\%$ in comparison to the baseline method.
As mentioned above, soft labels decrease results confidence. Adding the BCE term $(\alpha_{1}.NBCE)$ to the active loss increases confidence mostly on the true positive labels (Fig. \ref{fig:result} e, g) and increased recall 22\%, precision 14\%, and dice 8\%. Further, the probability distribution of active loss with BCE term is visually closer to the ground truth.

\section{Discussion}
In this paper, we proposed a noise-resistance deep learning pipeline using soft labels and an active-passive learning loss. The proposed method primarily addresses the problem of training a CNN using partially annotated data. This simple yet effective method has improved the detection rate of BML. Furthermore, we have generated soft labels that conformed to anatomic structures by combining features directly obtained from the images (pixel intensity and anatomy) with the ROI highlighting general areas of pathology, by using medical scores provided by human readers that can be used for semi-supervised learning. \par
The results using the OAI dataset for BML segmentation show that using a noise-resistant loss in combination with soft labels improves performance in both detection and segmentation tasks, compared to using noise-sensitive losses like CE or noise-robust losses such as MAE, SCE, or APL. Moreover, the proposed combination of active and passive losses for APL improved sensitivity on labeled areas without additional punishment for missed-labeled regions.
\subsection{Effect of segmentation loss on detection task}
Changes in segmentation loss function improved the result of the segmentation task (Dice score) and detection results (recall, precision, and IoU). This improvement can be attributed to the multi-tasking nature of the Mask RCNN approach. Mask RCNN learning tasks aim to concurrently identify instances, classify instances, and segment instances. Multi-task learning is proven to improve learning efficiency and prediction accuracy compared with training separate models for each task \cite{goodfellow2016deep,kendall2018multi}. In multi-task learning, generic parameters are softly constrained. Moreover, during back-propagation, the loss of one task has an impact on the concurrent tasks as well. Therefore, in IMaskRCNN, adding noise resistance loss to the mask-head contributes to regularization in other tasks as well.
\subsection{Multi-label Data With Label Noise}
Most of the previous proposed methods are applicable only for a single-label multi-class segmentation problem, where each data example is assumed to have only one true label. However, most medical image applications require the segmentation of multiple labels, for example, some pixels could be associated with multiple true class labels. However, methods that are developed based on MaskRCNN are suitable for multi-label applications since in MaskRCNN the $L_{mask}$ is defined only on positive ROIs and other mask outputs do not contribute to the loss. This constraint leads to no competition among classes for generating masks, which is suitable for identifying lesions in tissues multi-label data with weak labels.
\subsection{Challenges and limitations of the data}
Using KIMRISS annotation for training a network is challenging since we can only obtain weak labels. The reason is that KIMRISS objectives has not been providing pixelwise BML annotation and exact BML volume. Thus, some discrepancy exists between raters in the exact location and size of the templates. Furthermore, ROIs obtained from KIMRISS are rectangular shapes and do not provide pixel or shape information. In addition, labeling uncertainty (uncertainty of disagreement and uncertainty of single-target label \cite{ju2022improving}), which is common in medical images annotations, introduces another noise to these partial labels. Due to varying thresholds between readers for defining positive lesions, disagreement uncertainty exists for the KIMRISS annotation, which is measured as inter-observer reliability for the BML volume. Although the measured reliability suggests an acceptable confidence interval for clinical decisions \cite{jaremko2017validation}, these kinds of discrepancy affect model training. \par
Other sources of noise labeling were investigated through close observation and follow-up reading. The follow-up reading suggests that approximately $50\%$ of areas with BML were missed or underestimated in the consensus labels and more than $60\%$ of areas with BML had been underestimated (or missed) by single readers (mainly due to inattentional error and tunnel vision). It is possible to consider a part of these annotation errors as a random variable, since the same annotator may not make the same errors when annotating the same scan, a second time (after a period of time). Furthermore, our visual observation shows smaller or dimmer BML, BMLs in starting slices, and difficult slices were more likely to be missed by readers. \par
Moreover, we had much less data (only 1280 slices for training) compared to similar works who aimed to solve segmentation using weak label problem.
To mitigate the effect of low number of annotations, we can use self-supervised or knowledge distillation training. However, in this paper, our focus was only to investigate the effect of using soft-labeling and normalized APL loss.

\section{Conclusion and Future Work}
Combination of soft labeling and noise tolerant loss is an effective method for weakly supervised segmentation of medical images. It provides a convenient approach for improving the performance of DL models with minimal intervention to the existing methods, while revealing a novel way to design noise-robust loss functions for segmentation. The proposed method has the flexibility to quickly adapt to the state-of-the-art architectures and learning algorithms, unlike most of the current approaches that require changing the learning process to estimate correct labels of the training examples and learn under label noise. This method is suitable for knowledge distillation as it gives more information when compared to hard labels and is effective for one stage training. 
It does not suffer from class imbalance and, unlike U-net based architectures developed to be robust against noise, it is able to perform multi-class segmentation problem. \\
The research result is designed to integrate with the KIMRISS scoring online platform \cite{jaremko2017validation} for identifying BMLs and its volume in the clinical domain. Therefore, we need to investigate the effect of our method on network calibration and confidence in future work, and also measure the reader’s uncertainty.

\section{Acknowledgment}
Academic time for JJ, is made available by Medical Imaging Consultants (MIC), Edmonton, Canada. We thank the members of the OMERACT MRI in Arthritis Working Group for their participation and support in this project.





%
\bibliographystyle{IEEEtran}
\bibliography{IEEEabrv,myReferences}

\end{document}